# Human-Recognizable Robotic Gestures

John-John Cabibihan,[1,*] Wing-Chee So,[2] and Soumo Pramanik[1]

[1] *Social Robotics Laboratory, Interactive and Digital Media Institute; and the Department of Electrical and Computer Engineering, National University of Singapore*

[2] *Department of Educational Psychology, The Chinese University of Hong Kong*

[*]Corresponding author's contact: John-John Cabibihan (elecjj@nus.edu.sg)

The final publication is available at http://ieeexplore.ieee.org

## Abstract

For robots to be accommodated in human spaces and in humans' daily activities, robots should be able to understand messages from the human conversation partner. In the same light, humans must also understand the messages that are being communicated by robots, including the non-verbal ones. We conducted a web-based video study wherein participants gave interpretations on the iconic gestures and emblems that were produced by an anthropomorphic robot. Out of the 15 gestures presented, we found 6 robotic gestures that can be accurately recognized by the human observer. These were nodding, clapping, hugging, expressing anger, walking, and flying. We reviewed these gestures for their meaning from literatures in human and animal behavior. We conclude by discussing the possible implications of these gestures for the design of social robots that are aimed to have engaging interactions with humans.

# I. INTRODUCTION

There is a growing trend of robots that socially interact with people (e.g. [1-6]). Likewise, there are now robots that can assist in the provision of care and therapy to children, elderly, and atypical populations (e.g. [7-9]). Such kinds of robots are called social robots. Social robots are autonomous robots that are able to interact and communicate among themselves, with humans, and with the environment and are designed to operate according to the established social and cultural norms [10, 11].

For robots to achieve socially engaging interactions with humans, researchers have argued that robots are expected to learn and produce human-like attributes such as body movements [12-14]. Among different types of body movements, abundant research has shown that gestures accompanying speech can evoke meaningful social interaction. Thus, robotic gestures have become one of the key design features for engaging human-robot interaction [1].

So far, the majority of the works were on the development of robots that can understand human facial expressions, hand gestures, and body movements. For example, robots with sophisticated vision systems can now track the faces or movements of people. Using the captured data, several methods were proposed for the robot to classify human emotions or to follow the human teacher's instructions [15-20]. Another paradigm is the programming of robots by demonstration. In this approach, the robot captures the human demonstrator's motions through its on-board vision system [21]. The robot then processes the data through various machine learning algorithms and the gestures are replicated. Alternatively, the robot can also learn from the demonstrator's actions through the motion sensors that the demonstrator wears and the robot repeats the movements [22, 23]. To complement the initial data from the wearable sensors, some researchers have also taught robots by physically moving the robot's limbs to the desired positions or to correct the initial motion from the wearable sensors, i.e. kinesthetic teaching [24-27].

What remains to be addressed is whether human beings can understand the meaning of the head and body gestures that are produced by anthropomorphic robots. Research on the interpretation of robotic gestures, however, is relatively scarce. Among a few studies, Kanda et al. [28] found that human beings responded to body movements and utterances by a route guidance robot, while Oberman et al. [29] further reported that comprehending robotic actions might activate the mirror neuron system that was previously thought to be specifically selective for biological actions. However, little is known on whether humans can derive meaning from the gestures produced by social robots.

In this paper, we investigate various human-like gestures made by an anthropomorphic robot and we examine which among these gestures the human observer can recognize. Should a pattern of commonly recognizable robotic gestures emerge, robot designers and programmers can implement these to a robot with the likelihood that the human interaction partner can understand the gestures. This is relevant because the development of robots with lifelike appearance and behaviors require hardware and software systems that work in synchrony. Only after the hardware and software systems are completed can human-robot interaction experiments begin. All these take time and effort. Knowing which gestures to implement to the robot can reduce the development cycle time.

The primary focus of this paper is on iconic gestures and emblems (cf. [30]), whereby the gestures carry the semantic meaning. We presented participants with iconic gestures and emblems that were performed by a human and a robot demonstrator. The participants were then asked to identify the meaning of those gestures. We examined whether human beings can interpret the robot gestures in the same way as the human gestures.

The next section gives a background on the communicative gestures that humans are able to do. Section III describes the robot and the procedures on how it was programmed with the gestures. Section IV describes our experimental procedure. Section V presents the results for the agreement rates and response time. The findings and implications are discussed in Section VI. The concluding remarks are given in Section VII.

## II. COMMUNICATIVE HUMAN GESTURES

Language is more than words. "As the tongue speaketh to the ear, so the gesture speaketh to the eye," Sir Francis Bacon once said [31]. Even simple body movements, like eye gaze and head nods, allow the fluid exchange in the roles of speaker and listener [32-35]. Interestingly, humans have been found to be very sensitive to these nonverbal cues [36]. Previous research has shown that among conversation participants, the appropriate nonverbal gestures play a key role in helping communicate intent, instruct, lead, and build rapport [37-39]. Furthermore, earlier research has shown that nonverbal signals from the face, voice, posture, gesture, interpersonal distance, and positioning have physiological effects on the other person, which have been found to be distinct from the effects of linguistic information [40].

Gestures are the spontaneous movements exhibited by speakers from all cultural and linguistic backgrounds as they are engaged in conversations [41-43]. From a speaker's perspective, gesture facilitates communication. Iconic gestures and emblems are often used to complement speech since they are meaningful hand configurations, and thus, are clearly communicative [44]. For example, a thumb that is pointing upward while the rest of the fingers are curled indicates approval; an index finger drawing an arc may indicate the motion of a long jump.

Gesture not only facilitates speech production, but also speech comprehension [45, 46]. It was suggested that speakers gesture to maximize the information conveyed to listeners [47]. Thus, when information crucial to communication is not conveyed in speech, gestures assist in conveying the information instead. For example, a speaker produces gestures to show directions to the listeners, because gesture is a better modality to convey spatial information.

## III. System Configuration

We used the upper-body of a 10 degrees-of-freedom (DOF) anthropomorphic robot for the experiments (Scout, Dr. Robot, Ottawa, Canada; see Fig. 1). The robot has 2 DOF in the neck, 2 DOF in each shoulder, 1 DOF on each elbow and 1 DOF on each wrist. Inspired by typical human gestures, we evaluated an initial set of 25 gestures in a pilot study [48]. Fifteen gestures were selected for this paper. These robot gestures range from simple movements requiring only 1 DOF such as nodding, to more complex gestures such as hugging, which can be accomplished with the 8 DOF of the arm motion. Due to the limited DOF by the robot, we only included gestures that involved the head and arms.

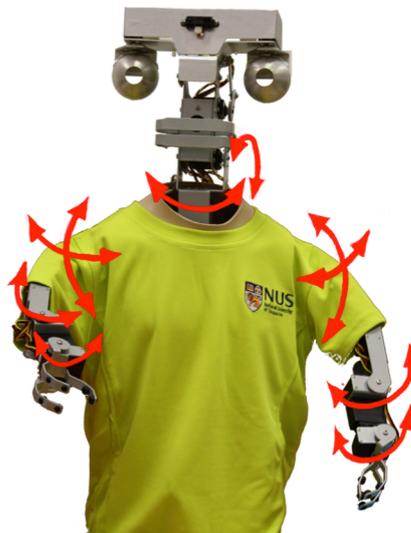

Fig. 1. The robot that was used to demonstrate the gestures. The arrows illustrate the 10 degrees of freedom of the robot.

For the human gesture to be replicated (Fig. 2a), a software application was developed to control the servomotors at the joints of the robot. Microsoft Visual Studio (version 2008) was used as the platform to develop the code in C# language, while Microsoft Robotics Developer Studio (MRDS; version R2) provided the environment for controlling the robot and for simulating the motion. As shown in Fig. 2b, a graphical user interface (GUI) was designed for simplifying the recording and playback of the gestures. Through the sliders in the GUI, the user can adjust the robot's head or arms to the desired positions. Each of the gestures was first broken down into multiple via points. These points are intermediate locations through which the robot joints can move. For each via point, the joint angles were defined using the GUI and the resulting sequence of joint angles was stored in the database (Fig. 2c).

The general configuration of the Scout robot matches that of a Lynx 6 Robotic Arm[1] (Lynxmotion, IL, USA). Because the MRDS supports the Decentralized System Service (DSS; Fig. 2d), we implemented the Lynx6Arm service, which was made available by Lynxmotion, for us to control the robot's movements. The Lynx6Arm service has a service partner named SSC32 service and this was used to send the commands to the servo-controller. At the time of gesture execution, the joint angles corresponding to the via points are passed in sequence as inputs to the DSS. The service uses inverse kinematics to determine the motor angles, which are required to achieve the corresponding joint angles of the robot. Finally, the servo controller (Fig. 2e) instructs the servomotors to move according to the calculated motor angles for the robot gesture to be performed (Fig. 2f).

---

[1] http://www.promrds.com/Chapter15/Lynx6Arm.htm

In addition, gesture execution speed was one of the important factors considered in mapping the human gestures to the robot. Research has shown that gesture speed influences emotional perception [49, 50], i.e., gestures expressing emotions of sadness are usually associated with slow movements whereas gestures expressing happiness are usually associated with faster movements. As the servomotors in the robot could not be operated at variable speed, varying the number of via points of a particular gesture controlled the speed of gesture execution. For example, gestures expressing sadness were typically modeled with approximately 200 via points whereas only 50 via points were used in the case of gestures for expressing happiness. During the execution of the gestures, these points were sent at a constant interval of 25 ms to the servo controller, implying that increasing the via points would produce a slower motion of the links. The resulting motions for each of the joints can then be recorded and can be played backed when necessary. Screen shots of the robot are shown in Fig. 3a.

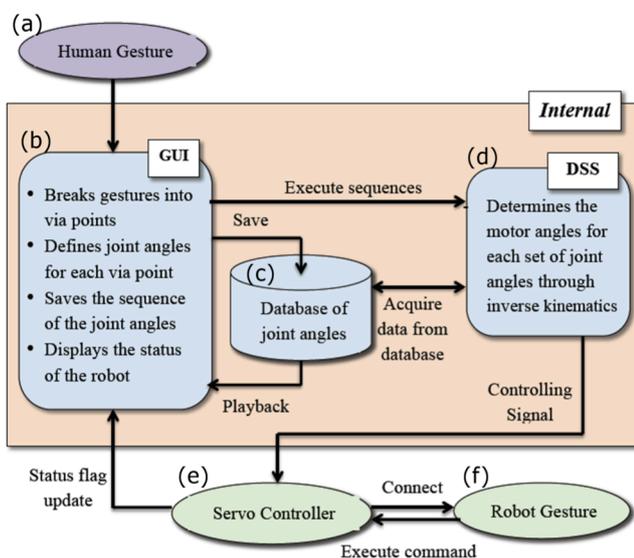

Fig. 2. Software framework for the iconic and emblematic robotic gestures.

## IV. Experiment

### A. Research Design and Participants

We designed an experiment in which we presented the participants a set of 15 gestures that both a human and a robot can perform. By having the human gesture as a reference, we can establish that a particular gesture is familiar and is thus recognizable by the viewer. Furthermore, we counterbalanced the presentation order such that one group of participants were presented the human gestures first, then the robot gestures, while another group of participants were shown the robot gestures, then the human gestures. We then compared whether the same gestures were recognized when a robot and a human produced them. We also compared the response latency across conditions. One hundred and twenty-two undergraduate and graduate students (70 males, 52 females, all 18-30 years old) from the National University of Singapore (NUS) were recruited by email. Of all participants, 65 participants were presented with robot gestures, followed by human gestures (Robot-Human, RH, condition); 57 students were presented with human gestures, followed by robot gestures (Human-Robot, HR, condition). Three units of a music player (iPod Shuffle, Apple, CA, USA) were raffled off to the participants.

## B. Human Gestures

A male demonstrator was selected to perform the gestures that the robot can do. To replicate the limited movements of the robot, the demonstrator was instructed not to use any body part other than his hands, arms, and neck while acting the gesture. He was instructed to keep a neutral expression and not to use any facial cues like blinking, smiling or frowning. The screen shots of the human gestures are shown in Fig. 3b. Please see the Appendix for the list of gestures and the corresponding interpretations that were given by the participants.

## C. Experimental Procedure

A web-based video study was conducted. The website was hosted on the NUS server. To ensure that an individual participates in the experiment only once, participants had to first register themselves with a valid email address. Because the presentation order was counterbalanced, two different versions of invitation emails were circulated for the Robot-Human[2] and the Human-Robot[3] groups. The invitation email contained the website's link to the experiment. The link opened the registration page that explained the objective of the experiment and that states the requirements to participate in the survey. A valid email address and the year of birth were required for registration. Upon successful registration, another email containing an activation link was sent to the participant's email address, which instructed the participant to click on the

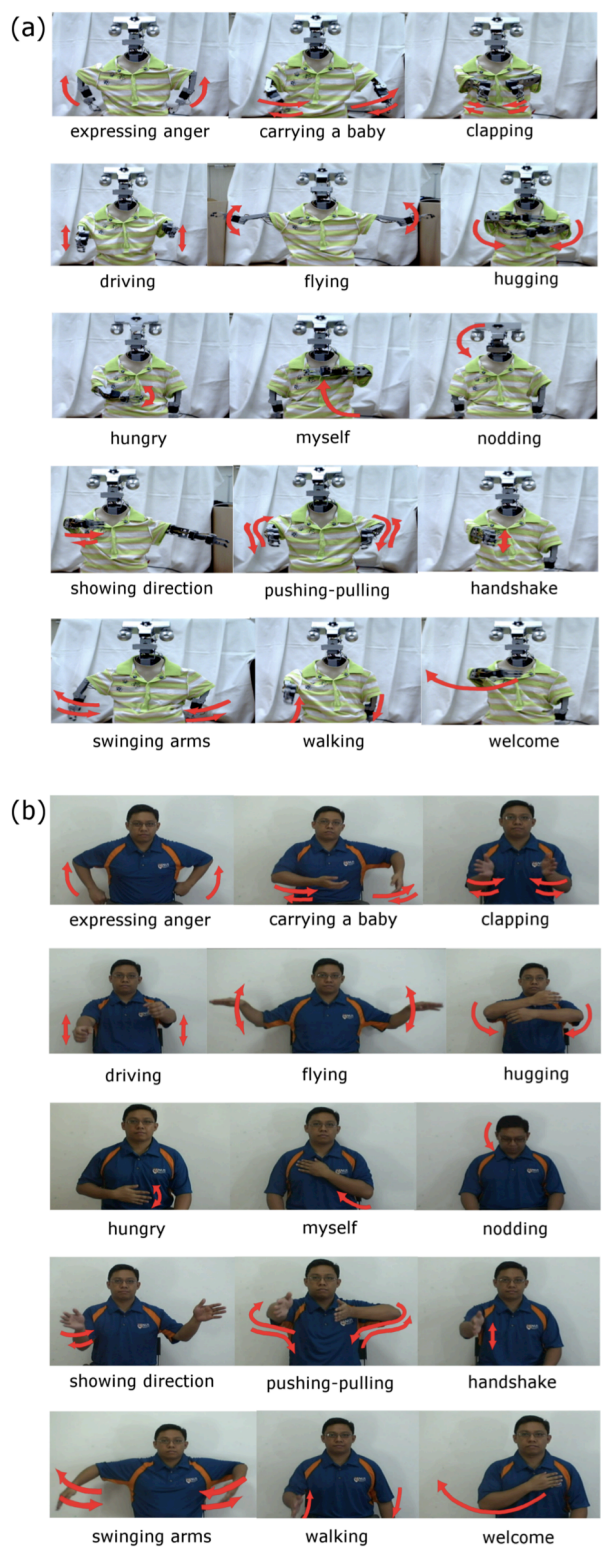

Fig. 3. Various gestures being performed (a) by a robot and (b) by a human demonstrator.

---

link to complete the activation process and proceed to the experiment. The webpage instructed the participants that each video will be played twice. Only then will they be able to enter their response. For gestures that they could not identify, participants were instructed to enter 'x' as their response.

The presentation sequence was played in accordance to the participant's assigned group. The participants were presented with the first set of 15 videos. Each gesture video lasts for 3 s and was repeated once. Then, they were asked to type the gesture that first came to their mind. To ensure that participants responded only with their first impression of the gesture, they were not allowed to replay the video and were given only 15 s to type the answer for each gesture. Our pilot tests showed that 15 s was the optimal time for a participant to type their response on the webpage. Response time was calculated as the time between the clicking of the 'Play' button for starting the video and the 'Next' button for proceeding to the subsequent gesture video. From this, 6 s was subtracted in order to account for the total running time of the video. Upon clicking the 'Next' button, the participant's interpretation of the gesture and the response time for the current video were stored in the database. To ensure that the earlier gestures did not influence the succeeding set of gestures, the gestures were presented in random order. The order of gestures in videos 1 to 10 was randomized separately from videos 11 to 15. This prevented two similar human or robot gestures to be shown close to one another.

### D. Coding

We analyzed the participants' interpretation of each gesture and identified the meaning of gesture that the majority of participants agreed with. Next, we counted the number of participants who agreed with the gesture's meaning (i.e., agreement rate). A 70% agreement rate indicates high similarity of responses among participants. This cut-off threshold was similar to the human gesture experiments in [51, 52]. A second coder then analyzed the similarity of the participants' interpretation of the gestures. The inter-rater similarity was 99.24% for the RH condition and 99.7% in the HR condition. All statistical analyses were performed using a commercial statistical package (IBM SPSS, version 19, NY, USA). A p-value of < 0.05 was considered statistically significant for analysis.

## V. RESULTS

Response latencies longer than 15 s were excluded from the analysis. Results with response time beyond 2.5 SDs of the sample mean were also excluded. This criterion removed 3.24% of the data in the Robot-Human condition and 4.71% in the Human-Robot condition.

### A. Agreement Rates

On average, the agreement rate for robot gestures and human gestures in the two conditions was 62.94% (SD = 33.88%). A mixed analysis of variance (ANOVA) with the type of gestures (robot, human) as the within-subject independent variable, condition (HR, RH) as the between-subject independent variable, and agreement rate as the dependent variable was conducted. We found a significant effect of gesture type, $F_{(1, 1323)} = 41.99$, $p < .001$, no effect of condition, $F_{(1, 1323)} = 1.46$, $p = ns$, and no interaction, $F_{(1, 1323)} = 2.88$, $p = ns$. Human gestures (M = 64%, SD = 48%) generally received higher agreement rate than robot gestures (M = 57%, SD = 50%).

Of all gestures, eight human gestures had mean agreement rates above 70%. These were clapping, nodding, walking, hugging, flying, expressing anger, shaking hands, and showing directions. Six robot gestures had mean agreement rates above 70%, notwithstanding the presentation order. These gestures were nodding, clapping, walking, hugging, flying, and expressing anger. More importantly, these robotic gestures were also well recognized in the pool of human gestures. Fig. 4 shows the mean agreement rates across different gestures.

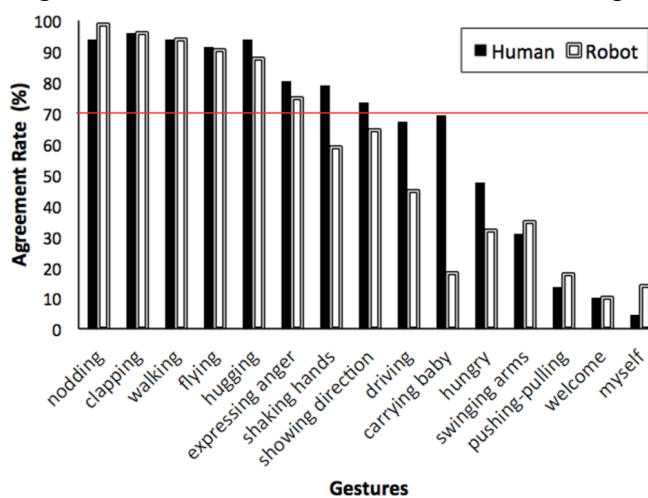

Fig. 4. The participants' responses for the mean agreement rates for the 15 gestures. The agreement rate denotes the agreement of the participant's responses to the gesture's meaning. The cut-off line represents the threshold agreement rate of 70%.

### B. Response Latencies

On average, participants spent 4.89 s (SD = 2.63 s) to respond to a gesture. A mixed ANOVA with the type of gestures (robot, human) as the within-subject independent variable, condition (HR, RH) as the between-subject independent variable, and response time as the dependent variable was conducted. We found a significant effect of condition, $F_{(1, 1323)} = 82.58$, $p < .001$, no effect of the gesture type, $F_{(1, 1323)} = 1.97$, $p$ = ns, and a significant interaction, $F_{(1, 1323)} = 197.27$, $p < .001$. In order to explore the interaction further, we conducted two separate paired-sample t-tests for the response time for both types of gestures in HR and RH conditions respectively. In the HR condition, participants responded longer to the human gestures (M = 4.90 s, SD = 2.53 s) than to the robot gestures (M = 3.88 s, SD = 1.81 s), $t_{(666)} = 9.85$, $p < .001$. In the RH condition, we found the opposite pattern. Participants responded longer to the robot gestures (M = 6.03 s, SD = 3.05 s) than to the human gestures (M = 4.78 s, SD = 2.52 s), $t_{(657)} = 10.05$, $p < .001$. Therefore, our findings showed that participants responded to the first set of gestures (either human or robot gestures) longer than to the second set of gestures.

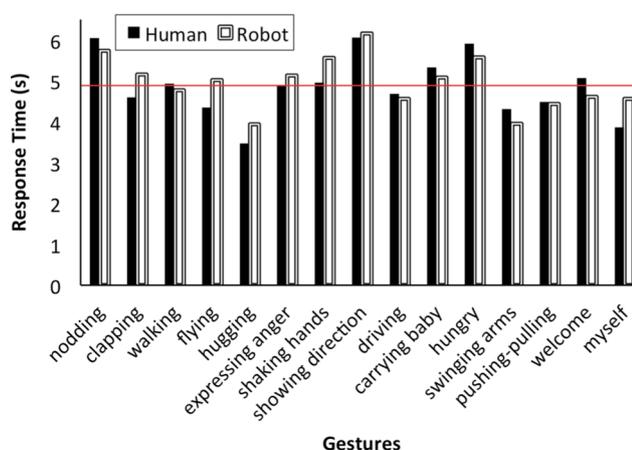

Fig. 5. The participants' mean response time for the 15 gestures. The cut-off line denotes the average time of 4.89 s in which participants responded to a gesture.

Among the human gestures, there were 8 gestures that had response times below the average response time of 4.86 s. These were hugging, the gesture to point to oneself (i.e.

myself), flying, swinging arms, pushing-pulling, clapping, driving, and expressing anger. Among the human gestures that had agreement rates higher than 70%, hugging, flying, clapping, and expressing anger can be recognized within 4.86 s. As for the mean response time of the other four gestures within the 70% agreement rate, walking, shaking hands, nodding, and showing directions can be recognized in about 6 s.

Among the robot gestures, there were 7 gestures that had mean response time below 4.86 s. These were swinging arms, hugging, pushing-pulling, driving, myself, welcome, and walking. Among these, hugging and walking had high agreement rates as shown earlier. The mean response times for the rest of the gestures were about 6 s or less. Fig. 5 shows the average response time for the different gestures.

## VI. DISCUSSION

We presented the participants with gestures that were being acted out by a human and a robot. Among the 15 gestures that the human demonstrator performed, the participants recognized 8 gestures, consisting of nodding, clapping, walking, hugging, flying, expressing anger, shaking hands and showing directions. In other words, there were approximately half of the gestures being recognized by human participants. Indeed, a 50% recognition rate, or thereabouts, is not uncommon in human gesture studies. In an earlier paper [52], experimental subjects were presented with 80 videotaped gestures and only 40 gestures (i.e. 50%) reached 70% of agreement rate. Note that the same cut-off rate was used in the current paper.

Among the 8 gestures that were recognized by the participants, six of the gestures were recognized by the participants when the robot performs those gestures. The gestures are: nodding, clapping, walking, hugging, flying, and the gesture for expressing anger. Other gestures were below the cut-off rate. The robotic gestures of shaking hands and showing directions—both using wrist motion and slight curling of the fingers—were less recognized presumably because the robot's hands have limited motion as compared to the hands of the human demonstrator. Regardless of the type of gestures, the participants were able to provide their responses within 5 s on average. They also responded faster to the set of gestures that were presented first.

We review the literatures in human and animal behavior to determine the meaning of the 6 gestures. We then discuss the possible implications of these gestures for the design of robots that are aimed to socially interact with humans.

### A. Gesture Meanings and Design Implications

*Nodding*

In human-to-human interactions, head movements can indicate agreement or affirmation, disapproval or negation, and many other semantic messages [37, 53-56]. It was observed that the nods by the listener in a conversation encouraged utterances of the speaker, which achieved an animated conversation between the speaker and the listener [57]. The recognition of nodding may have come easy as it is a primitive form of communication that has been observed in mother-infant interaction [58, 59].

When implemented on a sociable robotic penguin, it was discovered that the human interaction partner nodded more often when the robot nodded deliberately in response to the human's nods, as compared to when the robot did not nod [60]. Riek et al. [61] confirmed this co-nodding phenomenon through a robotic monkey's head that mimicked the nodding behavior of the human partner. They demonstrated that when the human nods, the robot nods in response. Interestingly, the human acknowledges the robot's nod. In a study on a route guiding robot, Kanda et al. [28] observed that people have generally rated their interactions with a robot highly when robots know when to nod and to gaze appropriately as compared with robots that have not learned these behaviors. Taken together, earlier research have shown that even with simple nodding gestures, robots have the ability to influence the interaction partner's behavior. Thus, nodding is a simple, yet meaningful gesture that should be implemented to a social robot.

*Clapping*

Humans clap with their hands as a sign of appreciation or approval. An audience normally expresses their appreciation for a good performance by the strength and length of their applause [62]. Clapping is one of human activities that have been studied in the field of video streaming analysis for sports, music and human-machine interaction, among many others [63-67]. Researchers have discovered that for apes in captivity, the apes clap in order to attract the attention of humans [68-70]; for apes in the wild, apes use clapping as a form of long distance communication to maintain group cohesiveness during instances of alarm [71, 72]. From these accounts in human and animal behavior, clapping appears to be a useful action for human-robot interaction.

Hanahara and Tada [73] proposed a new communication language based on clapping for humans to communicate better with robots. They described how the sound from clapping can be used to simplify how humans communicate with a robot in a manner that is analogous to a Morse-code type of message. They extended this idea further to develop a new clapping language that has syllables, words, and syntax of its own. We have yet to see examples in human-robot interaction wherein robots clap in order to show appreciation, approval or as a gesture to attract attention, similar to how humans and apes employ it.

*Hugging*

Emotions are better expressed through nonverbal behavior (e.g., [74, 75]). Among the nonverbal behaviors, social touching (e.g. hugs, patting, caress, handshake) appears to be an important modality [76]. Hugging or embrace is a form of affective touching that involves the clasping or clinging of one's arms to another person. It has been suggested that the warmth from a parent's touch may help children feel secure in their exploration of their environments [77, 78]. On the contrary, the lack of parental warmth can cause children to experience more stress (e.g., [79]). Recent medical evidence shows that hugging can have health benefits to help reduce blood pressure and increase levels of oxytocin. Oxytocin is the hormone that is involved in social bonding, the formation of trust and the increase of generosity in humans [80, 81]. Hugging has been recognized as an important component of parent-child interaction. This may explain why hugging is an easily recognizable gesture. A robotic nurse named Nancy was developed as a research platform to explore the effects of robotic gestures and social touching on humans [82]. Among the touching behaviors that have been investigated are hugging [83, 84], handshaking [85, 86], and patting [87].

*Expressing Anger*

In the current paper, the human and robot demonstrators adopted the body position in which the hands touched the hips while the elbows were flexed outwards. This gesture is also known as arms akimbo. The majority of the participants have associated this movement as a gesture to express anger. Anger is an emotional reaction when situations or circumstances are unfair or unjust, personal rights are not expected, or realistic expectations are not met [88]. Although anger is considered as a normal human emotion, the fields of robotics and virtual reality have been very active in trying to replicate this emotion and other emotional behaviors on robotic faces and body movements. Programming emotions into robots has been identified as an important step to create lifelike social robots for improving human-robot interaction [89]. For example, the Waseda Eye No.4 Refined II (WE4-RII) humanoid robot was designed to show human-like emotions [90]. Its robotic face has a total of 26 DOFs that can control the facial Action Units (cf. [91]) of the eyebrows, eyelids, eyes, mouth, and lips. When the authors compared the robot face and the human face for an angry facial gesture, the participants rated the emotional intensity of an angry human face significantly higher as compared to an angry robotic face (i.e. mean: human = 82% vs. robot = 60%; for more details on the experimental results, the interested reader is referred to Fig. 2 of ref. [90]). In other words, even with a high DOF robotic face, such as that of the WE4-RII, the participants rated the angry emotion from a human face to be higher as compared to when a robot expressed it.

Like the behavioral experimental results of WE4-RII, more participants in our experiment recognized the human angry gesture as compared to when the robot expressed it (human = 80% vs. robot = 75%; see Fig. 4). In contrast, the robot we used performed the arms akimbo gesture without additional facial movements. Due to the tight coupling of the robot's appearance and behavior, there are still many issues that have to be addressed in replicating human emotions through the facial expressions and full body gestures by robots [92]. Nonetheless, participants are able to recognize an angry gesture from a robot although the emotional intensity may not be as strong as when humans express it.

*Walking and Flying*

Walking has been extensively studied in the context of human motion recognition [63, 65, 67] and gait analysis [93-97] while the flapping gesture has not been often described in the literatures. Considering that robot companions will be expected to engage in playful interactions, one study investigated the possible full-body motions that a human playmate will do to a small humanoid robot [98]. Cooney et al. found that walking, flying, hugging and rocking-the-baby are some of the few gestures that emerged from the recorded interactions.

If robots were to be used for education and entertainment applications, storytelling was suggested to be a necessary skill [99]. Story telling has been found to be one of the most powerful tools to teach a new language to a child [100]. Thus, robots are now being used as teaching assistants in the classroom. Robots have been considered for the task because they can help enunciate foreign words that the teachers may find difficult. Moreover, robots have a special appeal to many children. The children's early acceptance of robotics technologies naturally leads to interactions that are fun and exciting. As a result, students can be more receptive to new knowledge. For example, iRobi is a home-based personal robot that teaches the English language [101] while Robovie teaches English vocabulary to Japanese children [102]. The commercially available RoboSapien (Wowwee, HK, China) was programmed to tell English stories to Taiwanese children in a classroom setting [103]. Story telling robots will benefit from action gestures, like walking or flying, to make their lexical meanings easily known. These behaviors will be especially useful to the beginning learners of foreign languages.

### B. Limitation of the Study

The robot that we used in this work will not be able to fully replicate the gestures that can be performed by humans due to the limited degrees of freedom that the robot can operate. Take the handshake gesture, for example (Please see Fig. 4). This gesture was recognized with close to 80% agreement rate when the human demonstrator made the gesture. When the robot demonstrated the same gesture, the agreement rate was just 60%. *A posteriori* analyses of the handshake videos show that there were movements in the human demonstrator's wrist and the fingers were oriented in a grasping pose as he was performing the handshake gesture. Additional movements at the robot's wrist and fingers were not possible due to the absence of actuators to perform a similar motion.

Nevertheless, it is important for us to know which among the robot's gestures can be perceived as similar to when the humans perform the gestures. We understand that a large number of DOFs on a robot causes it to be bulky, which can result in difficulties for practical implementation. In addition, more flexibility on the robotic arms requires an increase in the DOFs. As a drawback, increasing the DOFs has an effect on the robot's controllability and leads to higher costs. Nonetheless, a minimal set of robotic gestures has emerged. A more compact and cost effective robot will have strong implications for the toy industry, entertainment and educational robots, among others.

## VII. CONCLUSION

Communication is a two-way street. In other words, not only should robots understand messages from the human conversation partner, humans must also understand the messages that are being communicated by robots, including the non-verbal ones. In this study, we found six robotic gestures that can be accurately recognized by the human observer. Nodding and clapping are gestures that are common for acknowledgement or agreement; hugging and the angry gestures are gestures that express emotions while walking and flying are action gestures, which can be used for storytelling or teaching languages. The development cycle time for robot programming and testing can be reduced if roboticists know at the onset the basic robotic gestures that humans can understand. When programmed into robots, these gestures can lead to human-robot interactions that are natural, appropriate, and engaging.

ACKNOWLEDGMENT

The authors thank Wendy Yusson for the preparation of the robot videos and for the pilot experiments.

REFERENCES

[1]   T. Fong, I. Nourbakhsh, and K. Dautenhahn, "A survey of socially interactive robots," *Robotics and Autonomous Systems,* vol. 42, pp. 143-166, 2003.

[2]   C. Breazeal, "Toward sociable robots," *Robotics and Autonomous Systems,* vol. 42, pp. 167-175, 2003.

[3]   F. Tanaka, A. Cicourel, and J. R. Movellan, "Socialization between toddlers and robots at an early childhood education center," *Proceedings of the National Academy of Sciences of the United States of America,* vol. 104, pp. 17954-17958, 2007.

[4]   K. Severinson-Eklundh, A. Green, and H. Hüttenrauch, "Social and collaborative aspects of interaction with a service robot," *Robotics and Autonomous Systems,* vol. 42, pp. 223-234, 2003.

[5]   Y. Nakauchi and R. Simmons, "A social robot that stands in line," *Autonomous Robots,* vol. 12, pp. 313-324, 2002.

[6]   M. Malfaz, A. Castro-González, R. Barber, and M. A. Salichs, "A biologically inspired architecture for an autonomous and social robot," *IEEE Transactions on Autonomous Mental Development,* vol. 3, pp. 232-246, 2011.

[7]   T. Shibata and K. Wada, "Robot therapy: A new approach for mental healthcare of the elderly - A mini-review," *Gerontology,* vol. 57, pp. 378-386, 2011.

[8]   K. Dautenhahn, "Socially intelligent robots: Dimensions of human-robot interaction," *Philosophical Transactions of the Royal Society B: Biological Sciences,* vol. 362, pp. 679-704, 2007.

[9]   H. Kozima, C. Nakagawa, and Y. Yasuda, "Children-robot interaction: a pilot study in autism therapy," in *Progress in Brain Research.* vol. 164, 2007, pp. 385-400.

[10]  S. S. Ge, "Social robotics: Integrating advances in engineering and computer science (keynote speech)," in *Electrical Engineering/Electronics, Computer, Telecommunications and Information Technology (ECTI)* Mae Fah Luang University, Chang Rai, Thailand, 2007.

[11]  H. Li, J. J. Cabibihan, and Y. K. Tan, "Towards an effective design of social robots " *International Journal of Social Robotics,* vol. 3, 2011.

[12]  S. Coradeschi, H. Ishiguro, M. Asada, S. C. Shapiro, M. Thielscher, C. Breazeal, M. J. Mataric, and H. Ishida, "Human-Inspired Robots," *Intelligent Systems, IEEE,* vol. 21, pp. 74-85, 2006.

[13]  H. Ishiguro, "Android science: Conscious and subconscious recognition," *Connection Science,* vol. 18, pp. 319-332, 2006.

[14]  A. Cangelosi, G. Metta, G. Sagerer, S. Nolfi, C. Nehaniv, K. Fischer, J. Tani, T. Belpaeme, G. Sandini, F. Nori, L. Fadiga, B. Wrede, K. Rohlfing, E. Tuci, K. Dautenhahn, J. Saunders, and A. Zeschel, "Integration of action and language knowledge: A roadmap for developmental robotics," *IEEE Transactions on Autonomous Mental Development,* vol. 2, pp. 167-195, 2010.

[15]  Y. Sakagami, R. Watanabe, C. Aoyama, S. Matsunaga, N. Higaki, and K. Fujimura, "The intelligent ASIMO: System overview and integration," in *IEEE International Conference on Intelligent Robots and Systems,* 2002, pp. 2478-2483.

[16]  E. Sato, T. Yamaguchi, and F. Harashima, "Natural interface using pointing behavior for human-robot gestural interaction," *IEEE Transactions on Industrial Electronics,* vol. 54, pp. 1105-1112, 2007.

[17]  C. L. Sidner, C. Lee, C. D. Kidd, N. Lesh, and C. Rich, "Explorations in engagement for humans and robots," *Artificial Intelligence,* vol. 166, pp. 140-164, 2005.

[18]  S. Waldherr, R. Romero, and S. Thrun, "Gesture based interface for human-robot interaction," *Autonomous Robots,* vol. 9, pp. 151-173, 2000.

[19]  P. Andry, A. Blanchard, and P. Gaussier, "Using the rhythm of nonverbal human-robot interaction as a signal for learning," *IEEE Transactions on Autonomous Mental Development,* vol. 3, pp. 30-42, 2011.

[20]  A. Yorita and N. Kubota, "Cognitive development in partner robots for information support to elderly people," *IEEE Transactions on Autonomous Mental Development,* vol. 3, pp. 64-73, 2011.

[21]  C. Breazeal and B. Scassellati, "Robots that imitate humans," *Trends in Cognitive Sciences,* vol. 6, pp. 481-487, 2002.

[22] R. M. Voyles, J. D. Morrow, and P. K. Khosla, "Towards gesture-based programming: Shape from motion primordial learning of sensorimotor primitives," *Robotics and Autonomous Systems,* vol. 22, pp. 361-375, 1997.

[23] R. Dillmann, "Teaching and learning of robot tasks via observation of human performance," in *Robotics and Autonomous Systems.* vol. 47, 2004, pp. 109-116.

[24] D. Lee, C. Ott, and Y. Nakamura, "Mimetic communication model with compliant physical contact in human-humanoid interaction," *International Journal of Robotics Research,* vol. 29, pp. 1684-1704, 2010.

[25] M. Hersch, F. Guenter, S. Calinon, and A. Billard, "Dynamical system modulation for robot learning via kinesthetic demonstrations," *IEEE Transactions on Robotics,* vol. 24, pp. 1463-1467, 2008.

[26] S. Calinon and A. Billard, "Incremental learning of gestures by imitation in a humanoid robot," in *HRI 2007 - Proceedings of the 2007 ACM/IEEE Conference on Human-Robot Interaction - Robot as Team Member*, 2007, pp. 255-262.

[27] S. Calinon, F. D'Halluin, E. L. Sauser, D. G. Caldwell, and A. G. Billard, "Learning and reproduction of gestures by imitation," in *IEEE Robotics and Automation Magazine.* vol. 17, June 2010, pp. 44-54.

[28] T. Kanda, M. Kamasima, M. Imai, T. Ono, D. Sakamoto, H. Ishiguro, and Y. Anzai, "A humanoid robot that pretends to listen to route guidance from a human," *Autonomous Robots,* vol. 22, pp. 87-100, 2007.

[29] L. M. Oberman, J. P. McCleery, V. S. Ramachandran, and J. A. Pineda, "EEG evidence for mirror neuron activity during the observation of human and robot actions: Toward an analysis of the human qualities of interactive robots," *Neurocomputing,* vol. 70, pp. 2194-2203, 2007.

[30] D. McNeill, "Iconic Gestures of Children and Adults," *Semiotica,* vol. 62, pp. 107-128, 1986.

[31] F. Bacon, *The advancement of learning, Book 2.* London: Oxford University Press, 1891.

[32] S. Duncan, L. Brunner, and D. Fiske, "Strategy signals in face-to-face interaction," in *Journal of Personality and Social Psychology.* vol. 37, 1979, pp. 301-313.

[33] S. Duncan and G. Niederehe, "On signalling that it's your turn to speak," in *Journal of Experimental Social Psychology.* vol. 10, 1974, pp. 234-247.

[34] S. Duncan, "Some signals and rules for taking speaking turns in conversations," in *Journal of Personality and Social Psychology.* vol. 23, 1972, pp. 283-292.

[35] A. Kendon, "Some functions of gaze-direction in social interaction.," in *Acta psychologica.* vol. 26, 1967, pp. 22-63.

[36] D. J. Schneider, A. H. Hastorf, and P. C. Ellesworth, *Person Perception*, 2nd ed. Menlo Park, CA.: Addison-Wesley, 1979.

[37] L. Tickle-Degnen and R. Rosenthal, "The nature of rapport and its nonverbal correlates," *Psychological Inquiry,* vol. 1, pp. 285-293, 1990.

[38] M. Knapp and J. Hall, *Nonverbal Communication in Human Interaction*, 4th ed. Fort Worth, TX: Harcourt Brace College, 1997.

[39] J. K. Burgoon, "Nonverbal Signals," in *Handbook of Interpersonal Communication*, 2nd ed, M. L. Knapp and G. R. Miller, Eds. Thousand Oaks, CA: Sage, 1994, pp. 229-285.

[40] T. Kito and B. Lee, "Interpersonal perception in Japanese and British observers," *Perception,* vol. 33, pp. 957-974, 2004.

[41] W. Wundt, *Volkerpsychologie. Eine Untersuchung der Entwicklungsgesetze von Sprache Mythus und Zitte*, 4th ed. vol. 1, 1921.

[42] G. H. Mead, *Mind, Self and Society from the Standpoint of a Social Behaviorist*. Chicago: University of Chicago Press, 1934.

[43] P. Feyereisen and J. D. de Lannoy, *Gestures and Speech: Psychological Investigations*. New York: Cambridge University Press, 1991.

[44] R. M. Krauss, "Why do we gesture when we speak?," *Current Directions in Psychological Science,* vol. 7, pp. 54-59, 1998.

[45] A. Kendon, "Gesture and speech: How they interact," in *Nonverbal interaction*, J. M. Weimann and R. P. Harrison, Eds. Beverly Hills, CA: Sage, 1983.

[46] A. A. Cohen, "The communicative functions of hand illustrators," *Journal of Communication,* vol. 27, pp. 54-63, 1977.

[47] J. P. de Ruiter, "Can gesticulation help aphasic people speak, or rather, communicate?," *Advances Speech-Lang. Pathol.,* vol. 8, pp. 124-127, 2006.

[48] J. J. Cabibihan, W. Yusson, S. Salehi, and S. S. Ge, "Minimal Set of Recognizable Gestures for a 10 DOF Anthropomorphic Robot," in *Social Robotics*. vol. 6414, S. S. Ge, H. Li, J. J. Cabibihan, and Y. K. Tan, Eds.: Springer Berlin / Heidelberg, 2010, pp. 63-70.

[49] A. P. Atkinson, M. L. Tunstall, and W. H. Dittrich, "Evidence for distinct contributions of form and motion information to the recognition of emotions from body gestures," *Cognition,* vol. 104, pp. 59-72, 2007.

[50] F. E. Pollick, H. M. Paterson, A. Bruderlin, and A. J. Sanford, "Perceiving affect from arm movement," *Cognition,* vol. 82, pp. B51-B61, 2001.

[51] W. D. Goh, S. Lidia, M. J. Yap, and S. Hui Tan, "Distributional analyses in auditory lexical decision: Neighborhood density and word-frequency effects," *Psychonomic Bulletin and Review,* vol. 16, pp. 882-887, 2009.

[52] D. F. Yap, W. C. So, J. M. Yap, Y. Q. Tan, and R. L. Teoh, "Iconic gestures prime words," *Cognitive Science,* vol. 35, pp. 171-183, 2011.

[53] J. N. Bailenson, A. C. Beall, and J. Blascovich, "Gaze and task performance in shared virtual environments," *Journal of Visualization and Computer Animation,* vol. 13, pp. 313-320, 2002/// 2002.

[54] D. Heylen, "Head gestures, gaze and the principles of conversational structure," *International Journal of Humanoid Robotics,* vol. 3, pp. 241-267, 2006.

[55] A. Kendon, "Some Uses of the Head Shake," *Gesture,* vol. 2, pp. 147-182, 2002.

[56] E. Z. McClave, "Linguistic functions of head movements in the context of speech," in *Journal of Pragmatics*. vol. 32, 2000, pp. 855-878.

[57] J. D. Matarazzo, G. Saslow, A. N. Wiens, M. Weitman, and B. V. Allen, "Interviewer Head Nodding and Interviewee Speech Durations," *Psychotherapy: Theory, Research and Practice,* vol. 1, pp. 54-63, 1964.

[58] W. S. Condon and L. W. Sander, "Neonate movement is synchronized with adult speech: interactional participation and language acquisition," *Science,* vol. 183, pp. 99-101, 1974.

[59] N. Kobayashi, T. Ishii, and T. Watanabe, "Quantitative evaluation of infant behaviour and mother–infant interaction," *Early Development and Parenting,* vol. 1, pp. 23-31, 1992.

[60] C. L. Sidner, C. Lee, L. P. Morency, and C. Forlines, "The effect of head-nod recognition in human-robot conversation," in *HRI 2006: Proceedings of the 2006 ACM Conference on Human-Robot Interaction*, 2006, pp. 290-296.

[61] L. D. Riek, P. C. Paul, and P. Robinson, "When my robot smiles at me: Enabling human-robot rapport via real-time head gesture mimicry," *Journal on Multimodal User Interfaces,* vol. 3, pp. 99-108, 2010.

[62] Z. Neda, E. Ravasz, Y. Brechet, T. Vicsek, and A. L. Barabasi, "The sound of many hands clapping," *Nature,* vol. 403, pp. 849-850, 2000.

[63] A. Diaf, R. Benlamri, B. Boufama, and R. Ksantini, "A novel eigenspace-based method for human action recognition," in *2010 5th International Conference on Digital Information Management, ICDIM 2010*, 2010, pp. 182-187.

[64] H. G. Kim, S. Roeber, A. Samour, and T. Sikora, "Detection of goal event in soccer videos," in *Proceedings of SPIE - The International Society for Optical Engineering*, 2005, pp. 317-325.

[65] Part, S. Danafar, and N. Gheissari, "Action recognition for surveillance applications using optic flow and SVM," in *Lecture Notes in Computer Science*. vol. 4844 LNCS, 2007, pp. 457-466.

[66] S. Sempena, N. U. Maulidevi, and P. R. Aryan, "Human action recognition using Dynamic Time Warping," in *Proceedings of the 2011 International Conference on Electrical Engineering and Informatics, ICEEI 2011*, 2011.

[67] S. Singh and J. Wang, "Human activity recognition in videos: A systematic approach," in *Lecture Notes in Computer Science*. vol. 4224 LNCS, 2006, pp. 257-264.

[68] S. R. Poss, C. Kuhar, T. S. Stoinski, and W. D. Hopkins, "Differential use of attentional and visual communicative signaling by orangutans (Pongo pygmaeus) and gorillas (Gorilla gorilla) in response to the attentional status of a human," *American Journal of Primatology,* vol. 68, pp. 978-992, 2006.

[69] A. W. Fletcher, "Clapping in chimpanzees: Evidence of exclusive hand preference in a spontaneous, bimanual gesture," *American Journal of Primatology,* vol. 68, pp. 1081-1088, 2006.

[70] A. S. Pollick and F. B. M. De Waal, "Ape gestures and language evolution," *Proceedings of the National Academy of Sciences of the United States of America,* vol. 104, pp. 8184-8189, 2007.

[71] A. K. Kalan and H. J. Rainey, "Hand-clapping as a communicative gesture by wild female swamp gorillas," *Primates,* vol. 50, pp. 273-275, 2009.

[72] J. M. Fay, "Hand-clapping in western low land gorillas," *Mammalia,* vol. 53, pp. 457-458, 1989.

[73] K. Hanahara and Y. Tada, "Human-robot communication with hand-clapping language," *Journal of Computers,* vol. 3, pp. 58-66, 2008.

[74] P. Ekman, "Lying and nonverbal behavior: Theoretical issues and new findings," *Journal of Nonverbal Behavior,* vol. 12, pp. 163-175, 1988.

[75] A. W. Siegman and S. Feldstein, *Nonverbal behavior and communication.* Hillsdale, NJ: Erlbaum, 1987.

[76] J. J. Cabibihan, S. Pattofatto, M. Jomaa, A. Benallal, and M. C. Carrozza, "Towards Humanlike Social Touch for Sociable Robotics and Prosthetics: Comparisons on the Compliance, Conformance and Hysteresis of Synthetic and Human Fingertip Skins," *International Journal of Social Robotics* vol. 1, pp. 29-40, 2009.

[77] J. S. Carton and E. E. R. Carton, "Nonverbal maternal warmth and children's locus of control of reinforcement," *Journal of Nonverbal Behavior,* vol. 22, pp. 77-86, 1998.

[78] J. S. Carton and S. Nowicki, Jr., "Antecedents of individual differences in locus of control of reinforcement: A critical review," *Genetic, Social, and General Psychology Monographs,* vol. 120, pp. 31-81, 1994.

[79] J. S. Carton and S. Nowicki Jr, "Origins of Generalized Control Expectancies: Reported Child Stress and Observed Maternal Control and Warmth," *Journal of Social Psychology,* vol. 136, pp. 753-760, 1996.

[80] M. Kosfeld, M. Heinrichs, P. J. Zak, U. Fischbacher, and E. Fehr, "Oxytocin increases trust in humans," *Nature,* vol. 435, pp. 673-676, 2005.

[81] P. J. Zak, A. A. Stanton, and S. Ahmadi, "Oxytocin increases generosity in humans," *PLoS ONE,* vol. 2, 2007.

[82] S. S. Ge, J. J. Cabibihan, Z. Zhang, Y. Li, C. Meng, H. He, M. R. Safizadeh, Y. B. Li, and J. Yang, "Design and Development of Nancy, a Social Robot," in *Proc of the 8th Intl Conf on Ubiquitous Robots and Ambient Intelligence*, Incheon, Korea, 2011.

[83] Y. Li, S. S. Ge, C. Yang, and K. P. Tee, "Model-free impedance control for safe human-robot interaction," in *Proc of IEEE Intl Conf on Robotics and Automation (ICRA)*, Shanghai, China, 2011, pp. 6021-6026.

[84] Y. Li, S. S. Ge, and C. Yang, "Impedance control for multi-point human-robot interaction," in *Proc of ASCC 2011 - 8th Asian Control Conference*, 2011, pp. 1187-1192.

[85] J. J. Cabibihan, R. Pradipta, and S. S. Ge, "Prosthetic finger phalanges with lifelike skin compliance for low-force social touching interactions," *Journal of NeuroEngineering and Rehabilitation,* vol. 8, p. 16, 2011.

[86] J. J. Cabibihan, R. Pradipta, Y. Z. Chew, and S. S. Ge, "Towards humanlike social touch for prosthetics and sociable robotics: Handshake experiments and finger phalange indentations," in *Lecture Notes in Computer Science* vol. 5744 LNCS, 2009, pp. 73-79.

[87] J. J. Cabibihan, I. Ahmed, and S. S. Ge, "Force and motion analyses of the human patting gesture for robotic social touching," in *Proc of IEEE Cybernetics and Intelligent Systems, Robotics, Automation and Mechatronics (CIS-RAM)*, Qingdao, China, 2011.

[88] S. L. Videbeck, *Psychiatric Mental Health Nursing*, 5th ed.: Lippincott Williams & Wilkins, 2010.

[89] J. Hirth, N. Schmitz, and K. Berns, "Towards social robots: Designing a emotion-based architecture," *International Journal of Social Robotics,* vol. 3, pp. 273-290, 2011.

[90] T. Chaminade, M. Zecca, S. J. Blakemore, A. Takanishi, C. D. Frith, S. Micera, P. Dario, G. Rizzolatti, V. Gallese, and M. A. Umilta "Brain response to a humanoid robot in areas implicated in the perception of human emotional gestures," *PLoS ONE,* vol. 5, 2010.

[91] P. Ekman, "Facial expression and emotion," *American Psychologist,* vol. 48, pp. 384-392, 1993.

[92] H. Ishiguro, "Scientific Issues Concerning Androids," *International Journal of Robotics Research,* vol. 26, pp. 105-117, 2007.

[93] S. Collins, A. Ruina, R. Tedrake, and M. Wisse, "Efficient bipedal robots based on passive-dynamic walkers," *Science,* vol. 307, pp. 1082-1085, 2005.

[94] M. M. Gross, E. A. Crane, and B. L. Fredrickson, "Effort-Shape and kinematic assessment of bodily expression of emotion during gait," *Human Movement Science,* 2011.

[95] M. P. Kadaba, H. K. Ramakrishnan, and M. E. Wootten, "Measurement of lower extremity kinematics during level walking," *Journal of Orthopaedic Research,* vol. 8, pp. 383-392, 1990.

[96] A. Koenig, X. Omlin, L. Zimmerli, M. Sapa, C. Krewer, M. Bolliger, F. Muller, and R. Riener, "Psychological state estimation from physiological recordings during robot-assisted gait rehabilitation," *Journal of rehabilitation research and development,* vol. 48, pp. 367-385, 2011.

[97] K. M. Naugle, J. Joyner, C. J. Hass, and C. M. Janelle, "Emotional influences on locomotor behavior," *Journal of Biomechanics,* vol. 43, pp. 3099-3103, 2011.

[98] M. D. Cooney, C. Becker-Asano, T. Kanda, A. Alissandrakis, and H. Ishiguro, "Full-body gesture recognition using inertial sensors for playful interaction with small humanoid robot," in *IEEE/RSJ 2010 International Conference on Intelligent Robots and Systems, IROS 2010 - Conference Proceedings*, 2010, pp. 2276-2282.

[99] B. Mutlu, J. Forlizzi, and J. Hodgins, "A storytelling robot: Modeling and evaluation of human-like gaze behavior," in *Proceedings of the 2006 6th IEEE-RAS International Conference on Humanoid Robots, HUMANOIDS*, 2006, pp. 518-523.

[100] C. A. Pesola, "Culture in the elementary foreign language classroom," *Foreign Language Annals,* vol. 24, pp. 331-346, 1991.

[101] J. Han, M. Jo, S. Park, and S. Kim, "The educational use of Home Robots for children," in *Proceedings - IEEE International Workshop on Robot and Human Interactive Communication*, 2005, pp. 378-383.

[102] T. Kanda, T. Hirano, D. Eaton, and H. Ishiguro, "Interactive robots as social partners and peer tutors for children : A field trial," *Human-Computer Interaction,* vol. 19, pp. 61-84, 2004.

[103] C. F. Shih, C. W. Chang, and G. D. Chen, "Robot as a storytelling partner in the english classroom - Preliminary discussion," in *Proceedings - The 7th IEEE International Conference on Advanced Learning Technologies, ICALT 2007*, 2007, pp. 678-682.

**John-John Cabibihan** was conferred with his PhD in biomedical robotics from the Scuola Superiore Sant'Anna, Pisa, Italy in 2007. At present, he is an Assistant Professor at the Department of Electrical and Computer Engineering of the National University of Singapore. Concurrently, he serves as the Deputy Director of the Social Robotics Laboratory; Associate Editor of the International Journal of Social Robotics; and the Chair of the IEEE Systems, Man and Cybernetics Society (Singapore Chapter; terms: 2011 and 2012). He was the Program Co-Chair of the 2010 International Conference on Social Robotics, Singapore, and Program Chair of the 2012 International Conference on Social Robotics at Chengdu, China. He is working on the core technologies towards lifelike touch and gestures for prosthetics and social robotics.

**Wing-Chee So** received her Ph.D. in the Department of Psychology at University of Chicago in 2007. Currently, she is the director of Language and Gesture lab and Assistant Professor in Department of Psychology at National University of Singapore. She is specialized in examining the role of gesture in communication and cognitive processes.

**Soumo Pramanik** is currently pursuing his M.Sc. in Mechatronics at National University of Singapore. His areas of interest include social robotics, machine vision and embedded systems.

# APPENDIX

| No. | Gesture | Related Words Given by the Participants |
|-----|---------|----------------------------------------|
| 1 | expressing anger | angry, annoyed, expression of anger, furious, angry emotion, hostility, irritation, showing anger |
| 2 | carrying a baby | baby, babysitting, carrying a baby, cradling a baby, cuddling, cuddling a baby, holding a baby, lullabye, making a baby to sleep, pacifying a baby by rocking, rock-a-bye-baby, rocking the baby, soothing a baby, swinging a baby |
| 3 | clapping | applause, approval, clap, clap happily, clapping, good job, good, hands clapping, happy, praise, well done |
| 4 | driving | drive, drive car, driving, driving a car, steering wheel of car, turning the steering wheel |
| 5 | flying | flap wings, flapping, flapping arms like wings, flapping hands like wings, flapping wings, fly, flying, imitating to fly, wing motion |
| 6 | hugging | embrace, hold, hug, hugging |
| 7 | hungry | feel your stomach hungry, hunger, hungry, stomach upset |
| 8 | myself | addressing yourself, me, myself |
| 9 | nodding | affirmative, agree, agreement, appreciation, approval, nod, nodding head, nodding in agreement, nodding in approval, nodding the head, nodding the head up and down, ok, saying yes |
| 10 | showing direction[4] | come, controlling traffic, directing, directing traffic, go, go ahead, go in, go left, go right, go to the right, keep right, move, move in that direction, move left, move on, move right, move there, move to my left, move to right, move to the right, move to this direction, regulate traffic, right side can pass, traffic, go to my right, go to your right, turn right, usher |
| 11 | pushing and pulling | churn, churning, mixing, push and pull, rubbing, turning something, washing hands |
| 12 | shake hands | handshake, shake, shake hand, shaking of hands |
| 13 | swinging arms | breakdance, dance step, dancing, dangling arms, gyration, robo-dance, robot dance |
| 14 | walking | jogging, march, running, swinging arm: walking, marching, walking slowly |
| 15 | welcome | hi, myself, welcome, welcoming |

[4]As for the directions, the participants did not specify whether the left or right directions were relative to the demonstrator or relative to the viewer

TABLE II
PARTICIPANT'S RESPONSES TO THE ROBOT GESTURES

| No. | Gesture | Related Words Given by the Participants |
|---|---|---|
| 1 | expressing anger | anger, angry, annoyed, displeasure, expressing anger, furious, irritation, losing patience, unhappiness |
| 2 | carrying a baby | carry baby, carrying a baby, cradling, lullaby, making a baby to sleep, rock, rocking a baby, rocking baby, rocking the baby |
| 3 | clapping | applause, clap, clap hands, clapping, good, happy |
| 4 | driving | drive, driving, driving a car, turning the steering wheel |
| 5 | flying | flap, flap wings, flapping, flapping arms, flapping wings, fly, flying, wing motion |
| 6 | hugging | a hug, embrace, hold, hug, hugging |
| 7 | hungry | hunger, hungry, hungry and sad |
| 8 | myself | choose me, gesturing to oneself, introducing itself, me, myself, point to oneself, self-introduction |
| 9 | nodding | affirmative, agree, agreement, appreciation, nod, nodding, yes |
| 10 | showing direction[4] | come this way, controlling traffic, directing, directing traffic, go, go ahead, go left, go right, go to the right, go to your right, keep right, maintaining crowd, move, move left, move right, move to right, move to the left, move to the right, move towards left, regulate traffic, signaling to turn left, traffic, turn right, usher, ushering |
| 11 | pushing and pulling | churning, pull and push, pulling and pushing a rope, push-pull |
| 12 | shake hands | handshake, handshaking, shake, shake hands, shaking of hands |
| 13 | swinging arms | break-dancing, dance move, dancing, dangling arms, hanging or dancing, loose hands, robo-dance, robot dance |
| 14 | walking | march, marching, run, running, walk, walking |
| 15 | welcome | open, peace, sincerity, welcome |